# Pathology Foundation Models

Mieko Ochi, Daisuke Komura, Shumpei Ishikawa

Department of Preventive Medicine, Graduate School of Medicine, The University of Tokyo

Corresponding Author's Information
Name: Shumpei Ishikawa
Address: 7-3-1 Hongo, Bunkyo-ku, Tokyo,113-0033 Japan
Telephone Number: 03-5841-3434
E-mail Address: ishum-prm@m.u-tokyo.ac.jp

Number of figures and tables
One figure and two tables.

Conflict of Interest:
None

Source of Funding:
This study was supported by the AMED Practical Research for Innovative Cancer Control under grant number JP 24ck0106873 and JP 24ck0106904 to S.I., JSPS KAKENHI Grant-in-Aid for Scientific Research (S) under grant number 2H04990 to S.I., and JSPS KAKENHI Grant-in-Aid for Scientific Research (B) under grant number 21H03836 to D.K.

IRB Approval Code and Name of the Institution
Not applicable.

Author Contributions
M.O. contributed to the search of previous publications and wrote the whole manuscript draft. D.K. and S.I. reviewed it critically.

Abstract

Pathology has played a crucial role in the diagnosis and evaluation of patient tissue samples obtained from surgeries and biopsies for many years. The advent of Whole Slide Scanners and the development of deep learning technologies have significantly advanced the field, leading to extensive research and development in pathology AI (Artificial Intelligence). These advancements have contributed to reducing the workload of pathologists and supporting decision-making in treatment plans. Recently, large-scale AI models known as Foundation Models (FMs), which are more accurate and applicable to a wide range of tasks compared to traditional AI, have emerged, and expanded their application scope in the healthcare field. Numerous FMs have been developed in pathology, and there are reported cases of their application in various tasks, such as disease diagnosis, rare cancer diagnosis, patient survival prognosis prediction, biomarker expression prediction, and the scoring of immunohistochemical expression intensity. However, several challenges remain for the clinical application of FMs, which healthcare professionals, as users, must be aware of. Research is ongoing to address these challenges. In the future, it is expected that the development of Generalist Medical AI, which integrates pathology FMs with FMs from other medical domains, will progress, leading to the effective utilization of AI in real clinical settings to promote precision and personalized medicine.

## 1. Introduction for digital pathology

*(a) Development of digital pathology*

For over 50 years, surgical pathology has played a vital role in diagnosing diseases, evaluating disease progression, and elucidating causes by observing patient tissue sections obtained from surgeries and biopsies through formalin fixation, embedding, and staining, which are then examined by trained pathologists under a microscope. During this process, several scoring systems, such as the Gleason score for prostate cancer and the Nottingham score for breast cancer, have been proposed to grade tumors based on their morphology. These scores provide essential information for determining treatment plans, but since they are determined based on the pathologist's subjective judgment, there is significant variability between pathologists. Moreover, some indicators impose a significant burden on pathologists in routine practice(1). However, the introduction of Whole Slide Scanners in the 1990s made it possible to create digital images of entire specimens with the same resolution as microscopes easily. This development has highlighted the application of image analysis and machine learning technologies to histopathology, leading to the rise of digital pathology, where the interpretation and analysis of digitized images (Whole Slide Images: WSI) can be performed quantitatively by computers. Furthermore, the remarkable speed of technological innovation in deep learning has led to the development of AI (Artificial Intelligence) that not only reduces the workload of pathologists but also aids in predicting patient prognosis and supporting treatment decisions based on WSI, resulting in research and development of AI with high potential clinical utility(2).

*(b) Current applications of AI in pathology*

The applications of current pathology AI can be broadly categorized into three types: 1) improving the accuracy and efficiency of pathology diagnosis, 2) predicting patient prognosis and supporting treatment decisions, and 3) integrating data with genomic information.

Some of 2)items include 3) as an elemental technology.

i) <u>Improving the accuracy and efficiency of pathology diagnosis</u>

AI enables accurate quantitative evaluation of various tissue features, such as immunohistochemical biomarker evaluation, cell counting, spatial arrangement of cells, structural density, distribution patterns, and tissue structure, allowing the development of diagnostic AI tools that perform diagnoses with the same or higher accuracy than general pathologists in specific tissues or provide information that pathologists find challenging to identify. Examples include diagnostic support tools that evaluate morphological features necessary for tumor assessment, such as tumor grade, histological type, and invasion extent, and automatic quantification tools for immunohistochemical staining intensity. For example, by using tools such as Mindpeak Breast Ki-67 RoI and Mindpeak ER/PR RoI, which automate the evaluation of Ki-67, estrogen receptor, and progesterone receptor in breast cancer, there have been reports of increased interobserver agreement among pathologists(3)。 Additionally, AI has been developed to optimize the pathology workflow by triaging case priorities, sorting cases to ensure urgent ones are reviewed first, and checking specimen quality and identification(4),(5),(6).

ii) <u>Predicting patient prognosis and supporting treatment decisions</u>

Several morphological features of histopathological tissues, such as tumor grade and tissue subtypes, established through years of research in pathology, have been shown to be useful as indicators for predicting patient prognosis. Similarly, various clinical and genetic information, such as treatment effects, responsiveness, resistance obtained from electronic medical records, and cancer gene panel test results, are also crucial for predicting patient prognosis. By using AI to effectively correlate and integrate these data with numerous histopathological findings obtained from pathology images, which have not been established as prognostic indicators before, studies have reported the development of more straightforward and accurate prognostic indicators. For example, Shi et al. developed a tumor microenvironment (TME) signature by automatically quantifying TME from pathology images in colorectal cancer, which aids in patient prognosis stratification(7).These studies have shown that by integrating individual histopathological features (and in some cases, clinical and genetic information) into a single classification system, it is possible to reflect the nature and behavior of tumors more accurately. Pathology AI applied to prognosis prediction is expected to contribute to the appropriate treatment selection and stratification of patients, promoting precision and personalized medicine(1), (2), (8).

iii) <u>Integrating data with genomic information</u>

There is growing attention to the research and development of AI that associates the phenotypes of histopathological tissue morphology with genomic profiles. For example, Jaume et al. developed biologically and histopathologically interpretable AI that integrates genomic information and WSI by using large datasets, such as TCGA(9), which contain both WSI and bulk transcriptome data(10). This research is essential for understanding the biological mechanisms that cause cancer and for selecting targeted therapies. Furthermore, studies have been conducted to estimate gene mutations and protein expression levels from pathology images to reduce the delay in patient treatment initiation due to the weeks-long time required for clinicians to obtain genetic and immunological test results(11),(12). Such research is progressing in the development of AI that integrates multimodal information and presents it in an interpretable form for pathologists and clinicians.

## 2. Introduction for foundation models

*(a) Development of Foundation Models*

With the advancement of social networking services (SNS), the global spread of COVID-19, which increased the amount of digital data, the improvement of computational efficiency due to hardware advancements, and the development of new AI architectures such as neural networks and transformers, large-scale AI models called Foundation Models (FMs), which are applicable to a wider range of tasks compared to traditional AI models, have emerged(13). The human-centered AI institute at Stanford defines an FM as "A foundation model is any model that is trained on broad data (generally using self-supervision at scale) that can be adapted (e.g., fine-tuned) to a wide range of downstream tasks."(14) Initially, large-scale language models trained on vast amounts of text data collected from the web were designed to solve various language-related tasks (information retrieval, text generation, sentiment analysis, chatbots, etc.) with high accuracy(15), (16). Subsequently, diverse FMs have been developed using various data modalities, such as images, audio, and point cloud data(17). According to the "AI Foundation Models: Update paper"(18) by the UK government's Competition and Markets Authority, updated in April 2024, with giant AI companies like Meta, Google, and OpenAI investing tens of millions of dollars to acquire developers. In the healthcare field, the number of papers related to medical foundation models has increased exponentially from 2018 to February 2024, reflecting the growing expectations and interest of healthcare professionals in the application of FMs in clinical practice. While there were only a few papers in 2018, more than 120 papers were published in 2023 alone(19).

*(b) Differences between Foundation Models and Non-Foundation Models*

The differences between FMs and non-FMs (Deep learning models) are summarized in Table 1(13). Generally, models have properties where the more complex the model, the more diverse relationships it can capture (expressiveness), and the more data, the better the model's performance (scalability)(20). FMs possess superior expressiveness and scalability based on larger models, training data, and parallelizable training methods. A common feature of both FMs and non-FMs is the need to obtain and utilize embeddings generated by the model when making inferences from input data. Embeddings are values representing the features of the input data as short vectors (small tensors), and the quality of these embeddings significantly impacts the accuracy of the model when applied to downstream tasks.

Table 1. Comparison between Foundation Models and Non-Foundation Models (Adapted from Table 1 in Reference [13])

| Characteristics | Foundation model | Non-foundation model (Deep learning model) |
|---|---|---|
| Model architexture | (Mainly) Transformer | Convolutional Neural Network |
| Model size | Vary large | Medium to large |
| Model applicable task | Many | Single |
| Performance on adapted tasks | State-of-the-art(SOTA) | High to SOTA |
| Performance on untrained tasks | Medium to high | low |
| Data Amount for model training | Very large | Medium to large |
| Using labeled data for model training? | No | Yes |

(c) Applications of Foundation Models in Healthcare

Foundation models (FMs) in healthcare can be broadly categorized into four types based on the data modality they handle. Here are some examples of their applications (downstream tasks)(19), (21):

i)      <u>Language Foundation Models (LFMs) for Natural Language Processing (NLP) in Healthcare</u>

Examples of use: Medical report generation, education for medical students/residents, patient self-diagnosis and mental health support.

ii)     <u>Vision Foundation Models (VFMs) for Image Data</u>

Examples of use: Image diagnosis and similar case retrieval, prognosis prediction for specific diseases, surgical assistance through detection of critical structures and lesions.

iii)    <u>Bioinformatics Foundation Models (BFMs) for Omics Data such as scRNA-seq, DNA, RNA, and Protein Data</u>

Examples of use: Sequence analysis, interaction analysis, structural and functional analysis, protein sequence generation, drug response and sensitivity prediction, disease risk prediction, and drug perturbation effect prediction.

iv)     <u>Multimodal Foundation Models (MFMs) integrating multiple modalities such as language, image, and bioinformatics</u>

Examples of use: Comprehensive diagnostic support based on multiple data modalities, medical image report generation, cross-search between text descriptions and chemical structures, promotion of research through dialogue with models harboring biological and medical knowledge, and providing advice on diagnosis and treatment based on patient inquiries and images.

## 3. Introduction for Histopathology Foundation Models

*(a) The Need for Foundation Models in Pathology*

Representative tasks in computational pathology include patient prognosis prediction, biomarker prediction, cancer and tissue subtype classification, cancer grading, diagnostic prediction, and immunohistochemical staining intensity scoring, as described in section 1(b). The development and clinical application of pathology FMs are desired for the following two reasons:

i)      <u>Annotation Cost</u>: Before the promotion of FM development, the approach to handle individual tasks involved creating supervised learning models based on data annotated by pathologists, which required substantial time and practical effort from pathologists for each disease, organ type, and task type(22). The average salary of a pathologist is $149 per hour (https://www.salary.com/research/salary/alternate/pathologist-hourly-wages), and assuming 5 minutes per slide, the annotation cost per pathology slide is approximately $12. The cost of annotation and the increase in working hours for pathologists can be a significant burden in the practice of clinical pathology(22), (23). Furthermore, the quality of annotation determines the performance of the trained model, requiring pathologists to conduct high-quality annotation, which adds to their burden and responsibility, becoming a bottleneck for model development for individual tasks. There is a demand for model development utilizing vast amounts of pathology image data with minimal or simple annotations per case.

ii)     <u>Lack of Public Datasets</u>: As of now, at least around 100 types of publicly available pathology image datasets(24) are accessible to model developers, many of which include hundreds to tens of thousands of WSIs. However, the diseases and organ types included in each dataset are limited, and the quality of the data varies. The largest public dataset including pathology images, TCGA, contains tens of thousands of WSIs but is limited to 32 types of cancers, making it impossible to cover the diverse diseases encountered in clinical practice.

*(b) Examples of Pathology Foundation Models*

As of June 2024, there are over 10 reports of foundation models specifically for pathology images. The origin and scale of datasets, the types of tissues included, and the learning methods vary for each FM. Many models are publicly available and usable for downstream tasks, though some are only available for limited use or are not publicly released at all (Table 2)(25), (26), (27), (28), (29), (30), (31), (22), (10), (32), (23), (33).

**Table 2. List of Pathology Foundation Models with Published Papers (including preprints) from October 2022 to June 2024.** Due to space constraints, not all models are included in this table. For a complete list of models, please refer to Supplementary Table 1.

\* When model name is not specified in the original paper, the first author's name is shown.

| Foundation Model Name | CTransPath | (Lunit)* | PLIP | Virchow | UNI | CONCH | PRISM | Prov-GigaPath | TANGLE | RudolphV |
|---|---|---|---|---|---|---|---|---|---|---|
| Publication Date | 2022 | 2022 | 2023 | 2024 | 2024 | 2024 | 2024 | 2024 | 2024 | 2024 |
| Journal Name | Medical Image Analysis | arxiv | Nature Medicine | Nature Medicine | Nature Medicine | Nature Medicine | arxiv | Nature | CVPR2024 | arxiv |

| Pretraining Dataset Name | TCGA PAIP | TCGA TULIP | OpenPath | MSKCC | Mass-100K Mass-1K Mass-22K | Educational sources PubMed Central Open Access Dataset | MSKCC | Dataset from Providence | TCGA TG-GATEs | Dataset from over 15 different laboratories across the EU and US TCGA |
|---|---|---|---|---|---|---|---|---|---|---|
| Number of GPUs/Type Used for Training | 48/ NVIDIA V100 GPUs | 64/ NVIDIA V100 GPUs | Not Specified | –/ NVIDIA A100 GPUs | 32/ NVIDIA A100 GPUs | 8/ NVIDIA A100 GPUs | 16/ NVIDIA V100 GPUs | 16 nodes × 4/ NVIDIA A100 GPUs | 8/ NVIDIA A100 GPUs | 16/ NVIDIA A100 GPUs |
| Embedding Level (patch/slide) | patch | patch | patch | patch | patch | patch | slide | slide | slide | patch |
| Number of WSIs | 29,763(TCGA) 2,457(PAIP) | 20,994(TCGA) 15,672(TULIP) | Not Specified | 1,488,550 | 100,426 | Not Specified | 587,196 | 171,189 | 2,074(TCGA) 6,597(TG-GATEs) | 133,998 |
| Number of Patch Images(M) | 15 | 32.6 | 0 | 2,000 | 100 | 1 | Not Specified | 1,300 | 15 | 1,200 |
| Number of Patients | Not specified | Not specified | Not specified | 119,629 | Not specified | Not specified | 195,344 | >30,000 | >1,864 | 34,103 |
| More than 10 types of organs in the dataset? | Yes | Yes | Yes | Yes | Yes | Yes | Yes | Yes | No | Yes |

| Staining Types (H&E/H&E+Others) | Not specified | H&E | H&E+Others | H&E | H&E | H&E+Others | H&E | H&E+Others | Not Specified | H&E+Others |
| --- | --- | --- | --- | --- | --- | --- | --- | --- | --- | --- |
| FFPE/frozen | FFPE, frozen | Not Specified | Not Specified | FFPE | FFPE | Not Specified | Not Specified | Not Specified | Not Specified | FFPE, Frozen |
| VFM/MFM | VFM | VFM | MFM | VFM | VFM | MFM | MFM | MFM | MFM | VFM |
| Model Publicly Available? | Yes | Yes | Yes | Yes | Yes | Yes | Yes | Yes | No | Yes |
| Reference Number | 33 | 25 | 28 | 27 | 29 | 30 | 31 | 22 | 10 | 32 |

In this paper, we provide an overview of these pathological FMs focusing on: *a) training datasets, modalities, and embedding levels, b) examples of downstream task usage, c) the status of public availability, and d) performance comparison between models.*

a) Training Datasets, Modalities, and Embedding Levels:
   i) <u>Dataset Origin and Scale</u>: Each model is pre-trained using public dataset (e.g., TCGA, PubMed Central Open Access Dataset (https://ncbi.nlm.nih.gov/pmc/tools/openftlist/), TG-GATEs(34)) and/or proprietary clinical pathological image dataset (collected from single or multiple institutions). For example, the Virchow(27) model utilizes a dataset of 1,488,550 WSIs from 119,629 patients stored at Memorial Sloan Kettering Cancer Center, currently the largest dataset of human pathological images.
   ii) <u>Organs and Tissue Types</u>: The types of organs and tissues covered by the datasets and their ratios are influenced by various factors, such as whether the dataset includes biopsy/surgical specimen WSIs, specific public datasets, or the case accumulation trends of the data collection medical facilities. While it is difficult to identify consistent trends, most datasets used to train FMs include images from more than ten different organs. Some models, such as Virchow, UNI(29), TANGLE(10), and RudolphV(32), are explicitly trained on datasets that include normal tissues, while it is not specified whether other models include only tumor datasets or also normal tissue images. There is no known experiment that examines the extent to which the inclusion of normal tissues alongside tumor tissues in FM training contributes to improved model performance in downstream tasks. Chen et al.(29) demonstrated that training on a combination of lab-derived datasets and a public dataset, which is composed of only normal tissue images, resulted in better performance compared to training on lab-derived datasets alone for the 43-class cancer type classification task. However, this improvement cannot be conclusively attributed to the mere increase in image count or the inclusion of normal tissue images.
   iii) <u>H&E vs. H&E + Other Stains</u>: While many pathological FMs are trained using datasets consisting solely of H&E (hematoxylin and eosin)-stained images, some models like CONCH and RudolphV also include immunostained and/or special stained images. CONCH conducted comparative experiments evaluating models trained with only H&E-stained images against those trained with other stain types included. The results showed that models incorporating various stains performed better on more tasks, such as tumor subtyping and grading, image-to-text retrieval, and text-to-image retrieval (8/13). However, in some classification tasks, models trained only on H&E-stained images either outperformed or showed only marginally lower performance compared to those with additional stains, indicating that incorporating diverse stains does not necessarily yield the best results.

iv) <u>VFM/MFM</u>: While many pathological FMs are VFMs, models like PLIP(28), CONCH, PRISM(31), Prov-GigaPath(22), and TANGLE(10) are MFMs. These models handle both pathological images and text (e.g., clinical pathology reports, social media, educational sources, and PubMed Central Open Access Dataset captions) or bulk transcriptome data in the case of TANGLE. These MFMs learn equivalence between paired image and other modality data, enabling the application of models to tasks like image-to-text retrieval, text-to-image retrieval, report generation, and gene expression analysis within images.

v) <u>Embedding Levels from Models</u>: WSIs are high-resolution virtual slide images of entire stained glass slide specimens, often exceeding several gigabytes in data size. WSIs have a pyramid structure with multiple layers of images at different magnifications, allowing for comprehensive observation of the entire specimen slide at any zoom level. Due to memory constraints, it is challenging to handle WSIs as whole images during model training, so they are typically divided into smaller patches (tiles) of a few hundred pixels. Therefore, some models output embeddings for each patch, while others provide WSI-level embeddings. Models like PRISM, Prov-GigaPath, and TANGLE output WSI-level embeddings, while others output patch-level embeddings. Aggregating patch-level embeddings to obtain WSI-level embeddings is possible, and the efficacy of each embedding level for different tasks remains debatable. Prov-GigaPath showed statistically significant improvements over previous patch-level embedding models across 16 tasks, including classification and image-to-text search. However, https://github.com/mahmoodlab/UNI indicates that UNI and CONCH outperformed Prov-GigaPath in four out of five classification tasks, including tumor grade classification and immunohistochemical protein expression intensity scoring.

b) Examples of Downstream Tasks

FMs are designed to be applicable to various downstream tasks, with appropriate accuracy verification conducted post-training. In healthcare, the tasks handled by VFMs/MEMs are as stated in section 2(c), but in this section, we focus on pathological images and list specific task contents by task category (Fig.1). The tasks mentioned here are based on those listed in FM papers(25), (26), (27), (28), (29), (30), (31), (22), (10), (32), (23), (33). We do not compare the performance of downstream tasks between models in this paper.

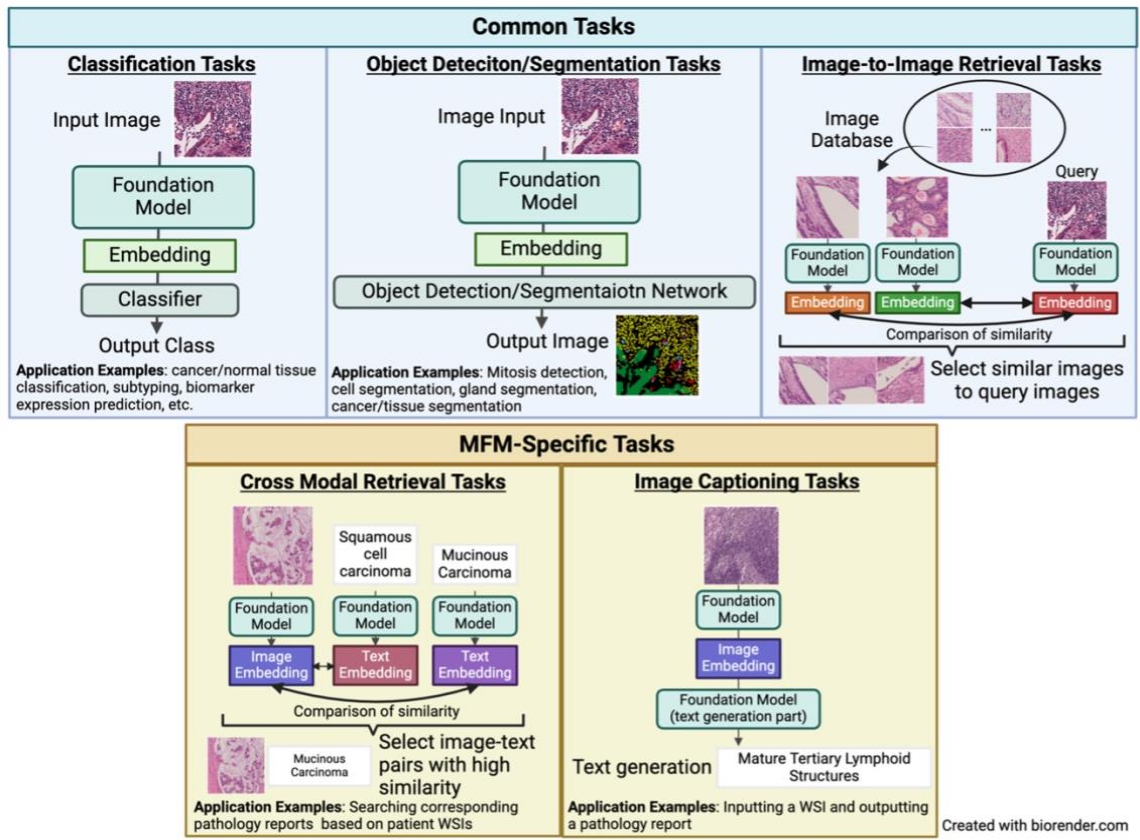

Figure 1. Task-specific schematic diagram of pathology foundation models.

Common Tasks: Tasks applicable to both VFMs and MFMs are primarily classification tasks, with fewer object detection, segmentation, and search tasks. VFMs like Lunit(25), UNI, RandolphV, and CTransPath(33) have been evaluated for these minor tasks. In several FM papers, the SegPath dataset, which we previously published, is used as a benchmark dataset for segmentation tasks. The SegPath dataset is a large-scale dataset consisting of over 1500 pairs of H&E-stained and immunohistochemically stained images(35).

i) <u>Classification Tasks:</u> Predicting which category a given image data (patch or WSI) belongs to. Examples: Disease detection, cancer detection, cancer/normal tissue classification, subtyping, disease diagnosis, rare cancer diagnosis, patient prognosis prediction, cancer recurrence prediction, cancer metastasis prediction, biomarker expression prediction (microsatellite instability, driver mutation, tumor mutation burden), treatment outcome prediction, immunohistochemical image protein expression intensity scoring.

ii) <u>Object Detection/Segmentation Tasks</u>: Detecting specific structures within images or predicting the category of each pixel. Examples: Mitosis detection, cell segmentation in H&E and immunohistochemical images, gland segmentation, cancer segmentation, tissue segmentation.

iii) <u>Image-to-Image Retrieval Tasks</u>: Searching for images with high similarity to a query image.

MFM-Specific Tasks:

i) <u>Cross Modal Retrieval Tasks</u>: Using an image or text query to search for corresponding text or image pairs. Examples: Retrieving corresponding pathology reports from a database based on patient WSIs.

ii) <u>Image Captioning Tasks</u>: Generating a summary text for an image. Examples: Inputting a WSI and outputting a pathology report.

*c) Status of Public Availability*

Most FMs are publicly accessible in an open-access format, allowing users to apply them to downstream tasks. However, some models (e.g., CONCH) have limited task availability compared to their original papers, or are not publicly available at all (e.g., Virchow, PRISM, RudolphV). Usage instructions for available models are typically described on their respective GitHub (A service for sharing and managing program source code online.) pages as referenced in the original papers.

*d) Performance Comparison Between Models*

Zeng et al.(36) conducted comparative accuracy experiments for CTransPath, UNI, Virchow, and Prov-GigaPath across 20 tasks in the domains of disease detection, biomarker prediction, and treatment outcome prediction. In the disease detection tasks, all four models achieved comparable performance. However, in the biomarker prediction and treatment outcome prediction tasks, UNI and Prov-GigaPath consistently demonstrated performance equal to or exceeding that of the other models. Specifically, for biomarker prediction in lung tissue, UNI and Prov-GigaPath outperformed the other models. The authors attributed these results to the higher representation of lung tissue in the pretraining datasets of UNI and Prov-GigaPath, suggesting that the proportion of relevant tissue in the training dataset might enhance the model's representational power for that tissue type.

4) Issues and Future Directions of Foundation Models

Thus far, we have outlined the development of digital pathology and FMs, their applications in healthcare, the introduction of pathological FMs, and specific examples of their use. In this section, we focus on the future directions of FM-related research and the potential issues users might face when employing FMs in practical AI applications in clinical settings.

(a) *Hardware Requirements*: Many FMs require specific hardware due to their immense complexity and scale. While these models demonstrate remarkable performance, their training and inference often necessitate substantial computational resources. For example, running Prov-GigaPath requires high-end GPUs such as the NVIDIA A100(37). The hardware investment needed for these models could be a challenge for resource-constrained medical institutions. Therefore, balancing model performance with hardware feasibility is a critical consideration when implementing these models clinically. Future research should aim to optimize these models to operate on less resource-intensive hardware, making them more accessible to a broader range of medical facilities.

(b) *Transition to Multimodal AI Assistants and Generalist Medical AI*: Currently, most developed pathological FMs are VFMs. However, various MFMs have been developed, focusing on combining language and visual modalities to produce pathology diagnostic reports from image inputs or focusing on molecular biology and visual modalities to identify significant genes in model-driven pathology diagnoses. These MFMs are becoming mainstream not only in pathology but across healthcare(19). Current self-supervised learning methods are not universally generalizable across all modalities and often need to be tailored to specific modalities(21). This limitation means integrating more modalities into FMs in real-world clinical settings is not yet feasible. In 2023, Lu et al.(38) developed a dialogue-based pathological AI assistant capable of prompt engineering, improving task accuracy by interactively providing appropriate questions and instructions. This assistant, built on previously developed pathological MFMs by their group, supports diagnostic assistance by describing histological findings and suggesting additional tests through interactions with pathologists. Recently, the concept of Generalist Medical Artificial Intelligence (GMAI)(39) has been proposed, utilizing vast and diverse data types, including images, electronic health records, test results, genomics, graphs, and medical texts, to perform various tasks across different medical specialties. The development of AI assistants integrating FMs from domains beyond pathology will likely drive comprehensive medical AI research and development towards realizing GMAI.

(c) *Hallucinations and Interpretability*: Models generate embeddings through probabilistic processes rather than truly understanding the logical meaning of the data(40). Consequently, FMs might generate information that does not actually exist (fabricated citations, information not inferable from input data) during generative tasks like pathology report generation (Hallucinations(41), (42)). Users must be aware of these potential hallucinations. Furthermore, the inference processes of these large-scale FMs, comprising hundreds of millions to trillions of parameters, are incredibly complex to understand, often described as the "black box" nature of AI. Unlike traditional medical devices that are typically more transparent and logical, this characteristic makes it difficult for people to trust the results of these algorithms(43). To address this, the field of "explainable AI(44)" is advancing, developing techniques to enhance human understanding of how AI models reach their outputs.

(d) Potential for Domain Shift and Information Updates: It is known that pathological specimens from overseas and Japan may show significant color and texture differences even with the same staining. Since all publicly available pathological FMs were developed using datasets from overseas facilities, they might not generalize well to Japanese specimens (WSIs) due to color and texture distribution differences (domain shift(45)). To address this issue, researchers have developed various methods for efficiently standardizing color tones and performing color augmentation. We have also previously created a large-scale pathological image dataset (PLISM)(46) that includes diverse tissues stained with H&E in various color tones to facilitate color augmentation. Additionally, the data used to train these large models, including PubMed and textbook knowledge, can become outdated quickly due to rapid advancements in medicine, necessitating regular updates to prevent inaccuracies. However, retraining these large-scale models with new data is costly. Researchers aim to improve these models' architectures using retrieval-augmented generation(15) (RAG), which allows models to reference external databases when generating responses, enhancing accuracy and explainability.

(e) *Concerns about Clinical Implementation of Foundation Models*: Despite the rapid development of FMs, their validation in real clinical settings remains insufficient, posing a barrier to clinical application. Like other AI models, FMs are currently regulated as medical devices by the U.S. Food and Drug Administration under a uniform software category developed for specific use cases(47). The World Health Organization released AI ethics and governance guidance for MFMs in 2024. It includes recommendations in the development and deployment for governments and developers(48). The guidance outlines over 40 recommendations for consideration by governments, technology companies, and health care providers to ensure the appropriate use of MFMs to promote and protect the health of populations. However, the current regulatory environment is inadequate to ensure the clinical safety and effective deployment of FMs. Among approved AI models, very few have been tested in randomized controlled trials, and none have involved FMs(49). The lack of transparency in trial reports and evaluated use cases is a significant issue. The Standard Protocol Items: Recommendations for Interventional Trials-AI and Consolidated Standards of Reporting Trials-AI guidelines(50), which were assessed by an international multi-stakeholder group from specific fields such as healthcare, statistics, computer science, law, and ethic in a two-stage Delphi survey and agreed upon in a two-day consensus meeting, have been formulated to improve standardization and transparency in clinical trials involving AI, but most published randomized controlled trials using AI technology have not strictly followed these established reporting standards(51). Users must be aware of the insufficient algorithm validation of FMs in real clinical settings.

## 5. Conclusion

With advancements in digital pathology, the utilization of Whole Slide Images (WSIs) has expanded, leading to active research and development of AI technologies based on them. Particularly, the emergence of FMs has broadened the scope of AI applications in medicine, allowing the development of AI models capable of handling various tasks by training on diverse data such as images, texts, and omics information. In the future, AI-based medical practice using FM-based AI assistants and GMAI is expected to be realized not only in pathology but also in real clinical settings, promoting efforts towards precision and personalized medicine. While medical AI is an extremely useful tool in practice, it is crucial for medical professionals, the primary users of FMs, to properly understand the effectiveness, considerations, and potential issues of using AI centered on FMs to ultimately benefit patients.

Supplementary Table 1. Complete list of pathology foundation models with published papers (including preprints) from October 2022 to June 2024. *When model name is not specified in the original paper, the first author's name is shown.

| Foundation Model Name | CTransPath | (Lunit)* | Phikon | PLIP | REMEDIS | Virchow | UNI | CONCH | PRISM | Prov-GigaPath | TANGLE | RudolphV |
|---|---|---|---|---|---|---|---|---|---|---|---|---|
| Publication Date | 2022 | 2022 | 2023 | 2023 | 2023 | 2023 | 2024 | 2024 | 2024 | 2024 | 2024 | 2024 |
| Journal Name | Medical Image Analysis | arxiv | arxiv | Nature Medicine | Nature biomedical engineering | arxiv | Nature Medicine | Nature Medicine | arxiv | Nature | CVPR2024 | arxiv |
| Pretraining Dataset | TCGA PAIP | TCGA TULIP | PanCancer40M TCGA-COAD | OpenPath | JFT-300M CAMELYON16 TCGA Colorectal tissue slides from the Institute of Pathology and the Biobank at the Medical University of Graz | Dataset from Memorial Sloan Kettering Cancer Center (MSKCC) | Mass-100K Mass-1K Mass-22K | Educational sources PubMed Central Open Access Dataset | Dataset from Memorial Sloan Kettering Cancer Center (MSKCC) | Dataset from Providence | TCGA TG-GATEs | Dataset from over 15 different laboratories across the EU and US TCGA |
| Number of GPUs/Type Used for Training | 48/ NVIDIA V100 GPUs | 64/ NVIDIA V100 GPUs | 16-64/ NVIDIA V100 GPUs | Not Specified | 16–256/ Google Cloud TPU cores | –/NVIDIA A100 GPUs | 32/ NVIDIA A100 GPUs | 8/ NVIDIA A100 GPUs | 16/ NVIDIA V100 GPUs | 16 nodes × 4/ NVIDIA A100 GPUs | 8/ NVIDIA A100 GPUs | 16/ NVIDIA A100 GPUs |
| Embedding Level (patch/slide) | patch | patch | patch | patch | patch | patch | patch | patch | slide | slide | slide | patch |
| Number of WSIs | 29,763(TCGA) 2,457(PAIP) | 20,994(TCGA) 15,672(TULIP) | 6,093(PanCancer40M) 441(TCGA-COAD) | Not Specified | 29,018(TCGA) | 1,488,550 | 100,426 | Not Specified | 587,196 | 171,189 | 2,074(TCGA) 6,597(TG-GATEs) | 133,998 |
| Number of Patch Images(M) | 15 | 32.6 | 43 | 0 | 50 | 2,000 | 100 | 1 | Not Specified | 1,300 | 15 | 1,200 |
| Number of Patients | Not specified | Not specified | 5,558 | Not specified | 10,705 | 119,629 | Not specified | Not specified | 195,344 | >30,000 | >1,864 | 34,103 |
| More than 10 types of organs in the dataset? | Yes | Yes | Yes | Yes | Yes | Yes | Yes | Yes | Yes | Yes | No | Yes |
| Staining Types (H&E/H&E+Others) | Not specified | H&E | H&E | H&E+Others | Not Specified | H&E | H&E | H&E+Others | H&E | H&E+Others | Not Specified | H&E+Others |
| FFPE/frozen | FFPE, frozen | Not Specified | FFPE | Not Specified | Not Specified | FFPE | FFPE | Not Specified | Not Specified | Not Specified | Not Specified | FFPE, Frozen |
| VFM/MFM | VFM | VFM | VFM | MFM | VFM | VFM | VFM | MFM | MFM | MFM | MFM | VFM |
| Model Publicly Available? | Yes | Yes | Yes | Yes | Yes | Yes | Yes | Yes | Yes | Yes | No | Yes |
| Reference Number | 33 | 25 | 26 | 28 | 23 | 27 | 29 | 30 | 31 | 22 | 10 | 32 |